\documentclass{article}
\usepackage{booktabs} 
\usepackage[style=numeric]{biblatex}
\usepackage{graphicx}
\usepackage
[
a4paper,
left=3cm,
right=3cm,
top=3cm,
bottom=3cm,
]{geometry}

\usepackage{hyperref}  
\usepackage{enumitem}

\usepackage[ruled]{algorithm2e} 

\SetAlFnt{\small}
\SetAlCapFnt{\small}
\SetAlCapNameFnt{\small}
\SetAlCapHSkip{0pt}
\IncMargin{-\parindent}

\newcommand{\dkh}[1]{\{#1\}}
\newcommand{\bigdkh}[1]{\big\{#1 \big\}}

\def\EE{\mathbb{E}}

\def\N{\mathcal{N}}

\def\goes{\rightarrow}

\def\A{\mathcal{A}}

\def\avail{\text{avail}}
\def\init{1}
\def\push{\text{sed}}

\def\given{\, | \,}

\newcommand{\indicator}[1]{ \mathds{1}_{\{#1\}}}

\def\EE{\mathbb{E}}

\def\BState{\State\hskip-\ALG@thistlm}

\def\transpose{\top}
\usepackage{amsmath,amssymb}
\usepackage{dsfont}
\usepackage{float}
\usepackage{xcolor}
\usepackage{multirow}
\usepackage{enumitem}

\usepackage{authblk}
\newcommand{\keywords}[1]{ \textbf{Key words: } #1}

\title{Personalized HeartSteps: A Reinforcement Learning Algorithm for Optimizing Physical Activity} 
	
\author{Peng Liao \thanks{Corresponding author. Email:\textit{pengliao@umich.edu}}}
\affil{Department of Statistics, University of Michigan}

\author{Kristjan Greenewald}
\affil{IBM Research}

\author{Predrag Klasnja}
\affil{School of Information, University of Michigan}

\author{Susan Murphy}
\affil{Department of Statistics, Harvard University}

\begin{document}
	
	\maketitle

	\begin{abstract} 
	With the recent evolution of mobile health technologies, health scientists are increasingly interested in developing just-in-time adaptive interventions (JITAIs), typically delivered via notification on mobile device and designed to help the user prevent negative health outcomes and promote the adoption and maintenance of healthy behaviors.  A JITAI involves a sequence of decision rules (e.g., a treatment policy) that takes the user's current context as input and specifies whether and what type of an intervention should be provided at the moment.  In this paper, we develop a Reinforcement Learning (RL) algorithm that continuously learns and improves the treatment policy embedded in the JITAI as the data is being collected from the user. This work is motivated by our collaboration on designing the RL algorithm in HeartSteps V2 based on data from HeartSteps V1. HeartSteps is  a physical activity mobile health application.  The RL algorithm developed in this paper is  being used in HeartSteps V2 to decide, five times per day, whether to deliver a context-tailored activity suggestion.   
	\end{abstract}

	
	\maketitle

	\section{Introduction}
	
	With the recent evolution of mobile health technologies, health scientists are increasingly interested in delivering interventions via notifications on mobile device at the moments when they can most readily help the user prevent negative health outcomes and promote the adoption and maintenance of healthy behaviors.  The type and timing of the mobile health interventions should ideally adapt to the real-time collected user's  context, e.g., the time of the day, the location, current activity and stress level.  This gives rise to the concept of a just-in-time adaptive intervention (JITAI) \cite{nahum2016just}.  Operationally, JITAI includes a sequence of decision rules (e.g., treatment policy) that takes the user's current context as input and specifies  whether and what type of an intervention should be provided at the moment. In practice,  behavioral theory along with expert opinion and analyses of existing data is often used to design the decision rules. However, these theories are often insufficiently mature  to precisely specify which particular intervention and when it should be delivered in order to ensure the interventions have the intended effects and optimize the long-term efficacy of the interventions.   As a result,  there is much interest in how best to use data to inform the design of JITAIs \cite{ghosh2017misspecified,tewari2017ads,bekiroglu2016control,rivera2018intensively,martin2018development, yom2017,paredes2014poptherapy,forman2018can,rabbi2015mybehavior,zhou2018personalizing}

	This paper develops a Reinforcement Learning (RL) algorithm to continuously learn, e.g., online, and optimize the treatment policy in the JITAI as  the user experiences the intervention.   This work is motivated by our collaboration on the design of the HeartSteps V2 clinical trial for individuals who have stage 1 hypertension.      In this clinical trial, the  HeartSteps V2 RL algorithm  learns whether to deliver a context-tailored physical activity suggestion as the trial progresses.   
	
	The remainder of the paper is organized as follows.  We first describe HeartSteps, including HeartSteps V1, and the current, in progress, clinical trial, HeartSteps V2. We then briefly review RL and identify  key challenges in applying RL to optimize JITAI treatment policies in mobile health. Existing mobile health studies that utilized RL are reviewed, as well as related RL algorithms. We then describe the proposed HeartSteps V2 RL algorithm, the implementation and an evaluation of this algorithm using a generative model built on HeartSteps V1 data. We discuss the performance of the proposed algorithm based on the initial pilot data from HeartSteps V2. We close with a discussion of  future work.
	\section{HeartSteps V1 and V2: Physical Activity Mobile Health Study}

HeartSteps V2 is an ongoing  90-day physical activity clinical trial for improving the physical activity of  individuals with blood pressure in the stage 1 hypertension range (120-130 systolic).	In this trial participants are provided a Fitbit tracker and a mobile phone application on the phone designed to help them improve their physical activity.  The participant first wears the Fibit tracker for  one week and then install the mobile app on the second week.  One of the interventions  is a contextually tailored physical activity suggestion that may be delivered at any of the five user-specified times during each day.  These five times are roughly separated by 2.5 hours, corresponding to the user's morning commute, mid-day, mid-afternoon, evening commute, and post-dinner times. The content of the suggestion is designed to encourage activity in the current context and thus the suggestions are intended to impact near time physical activity. The RL algorithm developed in this paper is being used to both decide at each time  whether to send the activity suggestion as well as to optimize these decisions. Currently HeartSteps V2 is being deployed in the field. We will provide an initial assessment of proposed algorithm in Section \ref{sec:real data}.
	
	In order to design HeartSteps V2,   our team conducted  HeartSteps V1, which is a 42-day physical activity study involving  37 healthy sedentary adults \cite{Liaoetal2015,KlasnjaAnnals, klasnja2015microrandomized, Dempsey_Significance}. In HeartSteps V1 whether to provide a  tailored activity suggestion  was randomized  at each of the 5 times per day with a constant probability of 0.30.  The data collected from HeartSteps V1  is used in this paper to (1) inform the design of RL algorithm for HeartSteps V2 (e.g., selecting the variables that are predictive of future step counts as well as the efficacy of the activity suggestion and  form a prior distribution) and (2) to create a simulation environment (e.g., the generative model) in order to evaluate the RL algorithm.  See sections \ref{sec: tuning parameters} and \ref{sec: simulation}.

	\begin{figure}
		\centering
		\includegraphics[width=0.8\linewidth]{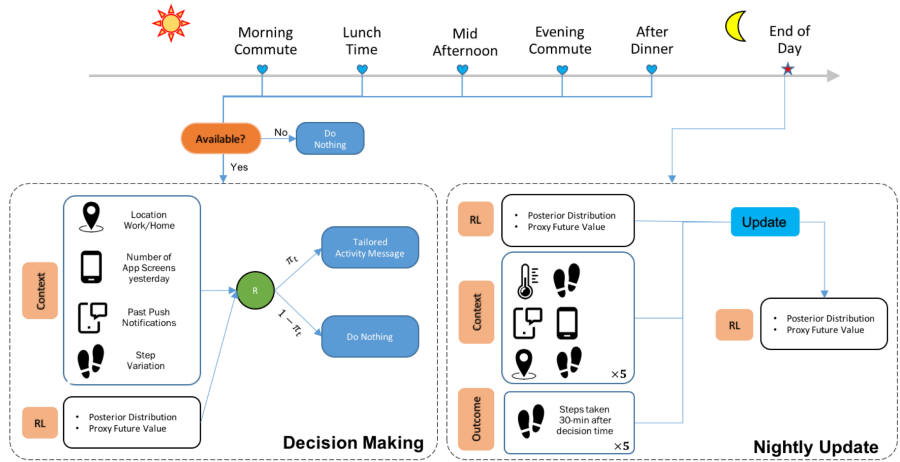}
		\caption{An illustration of HeartSteps V2.  }
		\label{fig:picture1}
	\end{figure}
	
	\section{Challenges to Applying RL in mHealth}
	
	\label{sec: challenges}
	Reinforcement Learning (RL) is an area of Machine Learning in which an algorithm learns how to act optimally by continuously interacting with the unknown environment \cite{sutton2018reinforcement}. The algorithm inputs the current state, selects the next action and receives the reward, with the goal of learning the best sequence of actions (i.e., the policy) to maximize the total rewards.  For example, in the case of HeartSteps,   the state is a set of features of the user's current and past context, the actions are whether to deliver an activity suggestion and the reward is a function of near time physical activity. A fundamental challenge in RL is the trade-off between exploitation (e.g.,  selecting the  action that appears best given data observed so far) and exploration (e.g., gathering information to learn the best action).  RL has seen rapid development in recent years and shown remarkable success across many fields, e.g., video games, chess-playing and robotic control.  However, many challenges remain that need to be carefully addressed before RL  can be usefully deployed to adapt and optimize  mobile health interventions. Below we discuss some of these challenges.

\begin{enumerate} [label=(C\arabic*)]
	 
		\item 	\label{challenge: delay}
		
		\textit{The RL algorithm must adjust for longer term effects of current actions.} In mobile health, interventions often tend to have positive effect on the immediate reward, but likely produce negative impact on the future rewards due to user habituation and/or burden \cite{Klasnja2008,dimitrijevic1972habituation}.   As such, the optimal treatment can only be identified by taking into account the impact of current action on the future rewards. This is akin to using a large discount rate (i.e., a long planning horizon) in RL.

		\item  	\label{challenge: learn fast}
		
		\textit{The RL algorithm should learn quickly and accommodate noisy data.}  Most  online RL algorithms  require the agent to interact many times with the environment prior to  performing well.  This is impractical in mobile health applications as users can lose interest and disengage quickly. Furthermore, because mobile health interventions are provided in uncontrolled, in situ complex environments 
		both context information as well as 
		rewards can be very noisy.  
		For example,  step count data collected from the wrist band is  noisy due to a variety of confounds including incidental hand movements.
		Additionally the sensors do not detect the entire context of the user;  non-sensed aspects of the current context act as sources of variance.   Such high noise settings typically requires even more interactions with the environment to select the optimal action.  Additionally, consideration of challenge \ref{challenge: delay} motivates a  long planning horizon. However, it has been shown that, in both practice and theory,  a discount rate close to 1 often lead to high variance and slow learning rates \cite{lehnert2018value,jiang2015dependence, arumugam2018mitigating,franccois2015discount}.   This results in the need to carefully trade off between bias and variance when designing the RL algorithm.

		\item  \label{challenge: model mis and nonstationary}
		\textit{The RL algorithm should accommodate some model mis-specification and non-stationarity.} Due to  the complexity of the context space 
		and unobserved aspects of the current context (e.g., engagement or burden),  the mapping from context to reward is likely to exhibit non-stationarity over longer periods of time.    Indeed, in the analysis of HeartSteps V1, there is strong evidence that the treatment effect of sending a activity suggestion on subsequent activity decreases with the time the user is in the study \cite{KlasnjaAnnals}, thus providing evidence of non-stationarity.

	\item  \label{challenge: data analysis}
	
	\textit{The RL algorithm should select actions so that after the study is over, secondary data analyses are feasible.}  This is particularly the case for experimental trials involving clinical populations.  In these settings, an interdisciplinary team is required to design the intervention and to conduct the clinical trial.   As a result multiple stakeholders will want to analyze the resulting data in a large variety of ways.   Thus, for example, off-policy learning \cite{thomas2016data,jiang2015doubly} and causal inference  \cite{boruvka2018assessing} as well as other more standard statistical analyses must be feasible after study end. 
	
\end{enumerate}

	\section{Existing RL-based Mobile Health Studies}
	
	There are few existing mobile health studies in which RL methods are applied to adapt the individual's intervention in real time. Here we only focus on the setting where the treatment policy is not pre-specified, but instead continuously learned and improved as more data is collected. 
	
	In \cite{yom2017}, a RL system was deployed to choose the different types of daily suggestions to encourage physical activity in patients with diabetes in a 26-week study. The authors uses a contextual bandit learning algorithm combined with a Softmax approach to select the  actions (daily suggestion) with the goal of maximizing increased minutes of activity.   Paredes et al. \cite{paredes2014poptherapy}  employed a contextual bandit learning algorithm combined with a Upper Confidence Bound approach to select among 10 types of stress management strategies when the participant requests an intervention in the mobile app with the goal of maximizing stress reduction.  A recent weight loss study is reported in \cite{forman2018can}, in which one of three types of interventions is chosen twice a week over 12-week period. Their RL system features an explicit separation between exploration and exploitation, e.g., 10 decision times is predetermined for exploration (e.g., randomly selecting the interventions at each decision time) and the rest of 14 decision times for exploitation (e.g., choosing the best intervention that maximizes the designed reward based on the history). In MyBehavior \cite{rabbi2015mybehavior}, a smartphone app that delivered personalized interventions for promoting physical activity and dietary health, used EXP3, a multi-arm bandit algorithm (e.g., context-free) to select the interventions.  While the RL methods in the aforementioned studies aim to select actions so as to optimize the immediate reward, in a recent physical study reported in \cite{zhou2018personalizing}, the RL system at the end of every week uses the participant's  historical daily step count data to estimate  dynamical system for the daily step count and use it to infer the optimal daily step goals for the next 7 days, with the goal to maximize the minimal number of step counts taken in the next week.   

	We argue that these RL algorithms are insufficient to address the challenges listed in Section \ref{sec: challenges} and thus require us to generalize these algorithms in several directions. First,  the above mentioned studies only use a pure data collection phase to initialize the RL algorithms;  however, often there are additional data from other participants, such as data from a  pilot study as well as prior expert knowledge.  Consideration of challenge \ref{challenge: learn fast} implies that it is critical to incorporate the prior knowledge  to speed up the learning in the early phase of the study. Second, the RL algorithms in these studies requires knowledge of the correct model for the reward  function, which is unlikely true due to the dimension and complexity of the context space and potential non-stationarity in challenge \ref{challenge: model mis and nonstationary}.   It is been empirically shown in RL literature that the performance of standard RL algorithms are quite sensitive to the model for the reward function  \cite{ghosh2017misspecified,dimakopoulou2017estimation,mintz2017non}.   Third, among the above mentioned studies, only the algorithm used in \cite{zhou2018personalizing} attempts to optimize rewards over a time period longer than the immediate time step.   It turns out that there is a bias-variance trade-off when designing how long into the future the RL should attempt to optimize rewards.  That is, only focusing on maximizing the immediate  rewards speeds the learning rate (e.g., due to lower estimation variance) compared with a full RL algorithm that attempts to maximize over a longer time horizon.   However, an RL algorithm focused on optimizing the immediate reward might end up sending too many treatments due to challenge \ref{challenge: delay}, i.e., the treatment tends to have a positive effect on immediate reward and negative effects on future rewards, and lead to poorer overall performance (akin to bias) than the algorithm that attempts to optimize over a longer time horizon to account for treatment burden and disengagement. Lastly, both \cite{paredes2014poptherapy} and  \cite{zhou2018personalizing} use algorithms that select the action deterministically based on the history, and  \cite{forman2018can} incorporate a pure exploitation phase.  It's known that action selection probabilities close to 0 or 1 cause the instability (i.e., high variance) in batch data analysis that use importance weights, e.g., in the off-policy evaluation \cite{thomas2016data,jiang2015doubly}.  This complicates challenge \ref{challenge: data analysis}.

	\section{Reinforcement Learning Algorithm in HeartSteps V2}
	
	In this section, we discuss the design of the RL algorithm in HeartStep V2; this algorithm determines whether to send the activity suggestion at each decision time.  
	We first give an overview of how the proposed algorithm addresses the challenges introduced in section~\ref{sec: challenges}.  Then we will specify each component in our setting, i.e., the decision times, action, states, and reward, and  formally introduce our proposed RL algorithm.

		\subsection{Addressing the Challenges}

	To address challenge \ref{challenge: delay}, we introduce a ``dosage'' variable based on the history of past treatments.  This is motivated by analyses of HeartSteps V1 in which moving to contexts with larger recent dosage appears to result in smaller immediate effect of treatment and lower future rewards.
A similar ``dosage'' variable was explored in a recent unpublished manuscript \cite{mintz2017non} where they developed a bandit algorithm, called ROGUE (Reducing or Gaining Unknown Efficacy) Bandits. They use the ``dosage'' idea to accomodate settings in which an (unknown) dosage variable causes  non-stationarity in the reward function.  Our use of dosage, on the other hand, is to form a proxy of the future rewards, in order to mimic a full RL setting (as opposed to the bandit setting) but managing variance in consideration of challenge \ref{challenge: learn fast}. We construct a proxy of the future rewards (proxy value) under a low dimensional proxy MDP model. Model-based RL  is well studied in the RL literature
\cite{osband2013more,fonteneau2013optimistic,ouyang2017learning}. In these papers, the algorithm uses a model for the transition function from  current state and action to next state.  Instead the proposed algorithm in this paper only uses the  MDP model to provide a low variance proxy to adjust for the longer term impact of actions on future rewards.

To further meet challenge \ref{challenge: learn fast}, a low-dimensional linear model  is used to model  differences in the reward function under alternate actions and as well as Thompson Sampling (TS). The use of a low-dimensional model is to trade off the bias and variance   to accelerate learning. TS is a general algorithmic idea that uses a Bayesian paradigm to trade-off between exploration and exploitation \cite{russo2018tutorial,russo2014learning}.  The use of TS allows us to incorporate prior knowledge in the algorithm through the use of a prior distribution on the parameters.  We propose using an informative prior distribution to speed up the learning in the early phase of the study as well as to reduce the variance and diminish the impact of noisy observation.   Note that TS-based algorithms have been shown to enjoy not only strong theoretical performance guarantees but strong empirical performance in many problems when compared to other state-of-the-art methods, such as Upper Confidence Bound \cite{kaufmann2012thompson,chapelle2011empirical,osband2016posterior,osband2017optimistic}.

To deal with  challenge \ref{challenge: model mis and nonstationary}, we use the idea of action-centering in modelling the reward. The motivation is to protect the RL algorithm from a misspecified model for  the ``baseline'' reward function (e.g., in HeartSteps example with binary actions, the baseline reward function is the expected number of future 30-min step count given the current state and no activity suggestion ). The idea of action-centering in RL was first explored in \cite{greenewald2017action} and recently improved in \cite{krishnamurthy2018semiparametric}. In both works, the RL algorithm is theoretically guaranteed to learn the optimal action  under no assumption about the baseline reward generating process (e.g., the baseline reward function can be non-stationary). However, neither of these methods attempts to reduce the noise in the reward. We generalize  action centering  for use in higher variance,  non-stationary reward settings.

Lastly, in consideration of challenge \ref{challenge: data analysis}, the actions in our proposed RL algorithm are selected stochastically via TS and furthermore we bound the TS probabilities away from 0 and 1  to  ensure the ability to conduct secondary analyses when the study is over.

	\subsection{Reinforcement Learning Framework}
\label{RLf}
Let the participant's longitudinal data recorded via mobile device be the sequence $$\{S_1, A_1, R_1, S_2, A_2, R_2, \dots, S_t, A_t, R_t, \cdots\}$$
Here $t$ indexes  decision time. In HeartSteps V1,  as in the planned HeartSteps V2, there are  five decision times each day. We also use $(l, d)$ to refer the $l$-th time decision time on study day $d$. For example, $(l,d) = (5, 3)$ refers to the 5-th time  in day 3, which corresponds to time $t = 5(d-1) + l = 15$.  $A_t \in \A$ is the action or treatment at  time $t$. The treatment is binary (i.e., the action space $\A = \{0, 1\}$), i.e., $A_t = 1$ if an activity suggestion is delivered and $A_t = 0$ otherwise. 
$R_{t}$ is the immediate reward collected after action $A_t$. In HeartSteps, the reward is the log transformation of the step count collected 30 minutes after the decision time. 
$S_t$ is the state vector at  decision time $t$. We decompose the state vector as $S_{t} = \{I_t, Z_t, X_t\}$. $I_t$ is used to indicate times at which only $A_t=0$ is  feasible and/or ethical. For example, if sensors indicate that the participant might be driving a car, then the suggestion should not be sent; that is, the participant is unavailable for treatment ($I_t = 0$).  $Z_t$ denotes features used to represent the current context at  time $t$.   In HeartSteps, these features  include current location, the prior 30-minute step count, yesterday's daily step count, the current temperature, as well as the measures of how active the participant has been around the current decision time over the last week. 
Lastly, $X_t \in \mathcal{X}$ is the ``dosage'' variable that captures our proxy for the treatment burden, defined based on the participant's treatment history. In contrast to HeartSteps V1, in HeartSteps V2, an additional intervention component, i.e., an anti-sedentary suggestion, will sometimes be delivered when the participant is sedentary.  As the anti-sedentary suggestion, in addition to the activity suggestions, can cause burden, it is included in defining the dosage variable.  Specifically, denote by $E_{t}$ the event that an activity suggestion is sent at decision  time $t-1$ (e.g., $A_{t-1} = 0$) and any anti-sedentary suggestion is sent between time $t-1$ and $t$. The dosage at the moment is constructed by first multiplying the previous dosage variable by $\lambda \in (0, 1)$ and incrementing it by 1 if any suggestions  were sent to the user since last  decision time. Specifically, starting with the initial value $X_1 = 0$, the dosage at time $t+1$ is defined as $X_{t+1} = \lambda X_{t} +  \mathds{1}_{E_{t+1}}$.  Based on the data analysis result from HeartSteps V1, we choose $\lambda = 0.95$; see section \ref{sec: tuning parameters} for how this value is selected.  

At each decision  time the RL algorithm selects the action based on each participant's current history (e.g., the past states, actions and rewards), with the goal to optimize the total rewards during the process. The proposed algorithm is stochastic, that is, the algorithm will output a probability to select an action.  
Denote the history up to the end of day $d$ by $H_d = \{S_{l, k}, A_{l, k}, R_{l, k} \}_{1\leq l \leq 5, 1 \leq k \leq d}$.  
The RL algorithm consists two components: (1)  the nightly update, e.g., $H_{d-1} \mapsto \{(\mu_d, \Sigma_d), \eta_d\}$  where $(\mu_d, \Sigma_d)$ denote parameters in  the posterior distribution for the reward and $\eta_d$ proxies the delayed effect on future rewards, both calculated at the end of the previous day $d-1$ (see below for more details), and (2) the  probability $\pi_{l, d}$, to select the action (e.g.,  $A_{l, d}$ is sampled from a Bernoulli distribution with probability $\pi_{l, d}$). Note that, at the beginning  of study (e.g., $d = 1$), both the distribution $(\mu_1, \Sigma_1)$ and the proxy of delayed effect $\eta_1$ are set based on the HeartSteps V1; see details in section \ref{sec: tuning parameters}. Throughout without loss of generality, we implicitly assume the probability $\pi_{l, d}$ is part of the state $S_{l, d}$.  The pseudo code of the proposed HeartSteps V2 RL algorithm is provided in Figure \ref{alg}.

\begin{algorithm}
	\DontPrintSemicolon 
	\KwIn{feature vectors $f(s)$ and $g(s)$, prior distributions $(\mu_{\alpha_0}, \Sigma_{\alpha_0})$ and $(\mu_{\beta}, \Sigma_{\beta})$, variance of noise $\sigma^2$. discount rate $\lambda$ in dosage, discount rate in proxy value $\gamma$, updating weight in proxy value $w$,   clipped probability $\epsilon_0$ and $\epsilon_1$. }
	
	\textbf{Initialize} $X_{1, 1} \leftarrow 0$, $\mu_1 \leftarrow \mu_{\beta}$, $\Sigma_1 \leftarrow \Sigma_{\beta}$\;
	\For{day $d = 1, 2, \dots, 90$} {
		\For{time slot $l = 1, 2, \dots, 5$} {
			Check the participant's availability $I_{l, d}$ \;
			Check event $E_{l ,d}$ and calculate $X_{l, d}$ based on the previous dosage and event $E_{l, d}$ \;
			Observe the context variable $Z_{l, d}$ \;
			Form the state, $S_{l, d} = \{I_{l, d}, Z_{l, d}, X_{l, d}  \}$\;
			\If{ available ($I_{l, d} = 1$)}{
				Calculate $\pi_{l, d}$   (\ref{action selection}), based on $\{(\mu_d, \Sigma_d), \eta_d \}$\;
				Sample $A_{l, d}$ from a Bernoulli distribution with probability $\pi_{l, d}$ \;
				Send the activity suggestion if $A_{l, d} = 1$. Otherwise,  do nothing \;
			}
			\Else{
				Do nothing
			}

		}
		Calculate the joint posterior distribution $\bar \mu_{d+1}, \bar \Sigma_{d+1}$:
		\begin{align*}
		& \bar \Sigma_{d+1} = \big(\frac{1}{\sigma^2}\sum_{k=1}^{d}\sum_{l=1}^{5} I_{l, k}\phi(S_{l, k}, A_{l, k}) + \bar \Sigma^{-1} \big)^{-1},  \\
		& \bar \mu_{d+1} = \bar \Sigma_{d+1}  \big(\frac{1}{\sigma^2}\sum_{k=1}^{d}\sum_{l=1}^{5} I_{l, k}\phi(S_{l, k}, A_{l, k})R_{l, k} + \bar \Sigma^{-1} \bar \mu\big) 
		\end{align*}
		\vspace{-3ex}  \;
		Set $\mu_{d+1} $ to the last $p$ elements of the $\bar \mu_{d+1}$ and $\Sigma_{d+1}$ to the bottom-right corner matrix of size $p$ by $p$ in $\bar \Sigma_{d+1}$\;
		Estimate the marginal reward function $r_1(x, a)$ and $r_0(x)$ and solve for the function $V^*$:
		\begin{align*}
		V(x, i) = \max_{a \in \A(i)} \bigdkh{ r_1(x, a) + \gamma \sum_{x', i'} \tau(x'|x, a) p_{\avail}^{i'} (1-p_{\avail})^{1-i'}V(x', i')}, ~ \forall (x, i)
		\end{align*}
		\vspace{-3ex} \;
		Calculate $H^* (x, a) = \sum_{x', i'} \tau(x'|x, a) p_{\avail}^{i'} (1-p_{\avail})^{1-i'}  V^*(x', i') $ and $H_{d+1} = (1-w) H_{\init} +  w H^*$\;
		Set $\eta_{d+1}(x) =  \gamma H_{d+1}(x, 0) - \gamma H_{d+1}(x, 1)$ for all $x$\;
	}

	\caption{HeartSteps V2 RL Algorithm}
	\label{alg}
\end{algorithm}

\subsection{Action Selection}

The reward function is given by $r_t(s,a)=\EE[R_{t} \given S_{t}=s, A_{t}=a, I_{t} = 1]$.   The  action selection developed here is based on a low dimensional linear model (challenge \ref{challenge: learn fast}) for the treatment effect:
\begin{align}
r_t(s,1) -r_t(s,0) =  f(s)^\transpose \beta \label{model:txt effect}
\end{align}
where  the feature vector, $f(s)$, is selected based on the domain science as well as on data analyses using HeartSteps V1; see section \ref{sec: tuning parameters} for the discussion of how the features are selected.
At the $l$-th decison time  on day $d$, availability is ascertained (i.e., $I_{l, d} = 1$). Then for $S_{l, d}=s$ with the dosage variable $X_{l, d}=x$, the action, $A_{l, d}=1$ is selected based on
\[
\Pr \bigdkh{f(s)^\transpose  \beta  >  \eta_d (x); ~\beta \sim \N(\mu_d, \Sigma_d)}
\]
where the random variable $\beta$, follows a Normal distribution $\N(\mu_d, \Sigma_d)$, e.g., the posterior distribution of the parameters, obtained at the end of previous day. 
The term $\eta_d (x)$ proxies the long-term, negative effect of delivering the activity suggestion at the moment given the current dosage level $X_{l, d} = x$ (see the detailed formulation of $\eta_d$ in section \ref{sec:proxy value}). Note that when $\eta_d (x) = 0$, we recover the bandit formulation, e.g., the action is selected to maximize the immediate rewards, ignoring any impact on the future rewards. 
The probability of sending an activity suggestion, $\pi_{l, d}$ (for $I_{l, d} = 1$, $S_{l, d}=s$, $X_{l, d}=x$) is a clipped version:
\begin{align}
\pi_{l, d}= \phi\left(\Pr \bigdkh{f(s)^\transpose  \beta  >  \eta_d (x); ~\beta \sim \N(\mu_d, \Sigma_d)}\right). \label{action selection}
\end{align}
The clipping function is  $\phi(\pi) = \min(1-\epsilon_0, \max(\pi, \epsilon_1)) \in [\epsilon_1, 1-\epsilon_0]$. This restricts the randomization probability of sending nothing and of sending an activity suggestion to be at least $\epsilon_0$ and $\epsilon_1$, respectively.  The probability clipping enables off-policy data analyses after the study is over (challenge \ref{challenge: data analysis}) and, furthermore, ensures that the RL algorithm will continue to explore and learn, instead of locking itself into a particular policy (challenge \ref{challenge: model mis and nonstationary}). In HeartSteps V2, $\epsilon_0 = 0.2$  and $\epsilon_1 = 0.1$.

\subsection{Nightly Updates}

The posterior distribution of $\beta$ for the immediate treatment effect and the proxy for the delayed effect are updated at the end of each day. Operationally, the nightly update is a mapping: $\{S_{l, k}, A_{l, k}, R_{l, k} \}_{1\leq l \leq 5, 1 \leq k \leq d}  = H_d     \mapsto \{(\mu_{d+1}, \Sigma_{d+1}), \eta_{d+1}\}$, that takes the current history up to day $d$ as the input and outputs the posterior distribution and proxy of delayed effect, which are used in the action selection in the following day (i.e., during day $d+1$).  We discuss each of them in turn.

\subsubsection{Posterior Update of Immediate Treatment Effect}

We use the following linear Bayesian regression ``working model'' for the reward to derive the posterior distribution for the treatment effect:
\begin{align}
& R_{t} =  g(S_t)^\transpose \alpha_0  +  \pi_t f(S_t)^\transpose \alpha_1 + (A_t - \pi_t) f(S_t)^\transpose \beta + \N(0, \sigma^2), ~\text{if }  I_t = 1 \label{model:rwrd}
\end{align}	
so that the working model for the reward function is $r_t(s,a)= g(s)^\transpose \alpha_0  +  \pi_t f(s)^\transpose \alpha_1 + (a- \pi_t) f(s)^\transpose \beta$.  The baseline feature vector $g(s)$ is used to approximate the baseline reward function: 
\begin{align}
& r_t(s,0) \approx g(s)^\transpose \alpha. \label{model:baseline rwrd}
\end{align}		
The baseline feature vector $g(s)$ is selected based on the domain science and data analyses using HeartSteps V1; see section \ref{sec: tuning parameters} for a discussion.	 The use of 	$\pi_t$ in (\ref{model:rwrd}) is unusual but provides a number of advantages as follows.	
Consider the action-centered term, $(A_t - \pi_t)$, in the working model (\ref{model:rwrd}).  As long as the  treatment effect model (\ref{model:txt effect}) is correctly specified, the estimator of $\beta$ based on the model (\ref{model:rwrd}) is guaranteed to be unbiased even when the baseline reward model (\ref{model:baseline rwrd}) is incorrect  \cite{boruvka2018assessing}, for example, due to the non-linearity in $g(s)$ or non-stationarity ($\alpha$ changes over time). That is, through the use of action centering, we achieve the robustness against mis-specification of the approximate baseline model, (\ref{model:baseline rwrd}), addressing challenge \ref{challenge: model mis and nonstationary}.     The rationale of including the term $\pi_t  f(S_t)$ in the Bayesian regression working model (\ref{model:rwrd}) is to capture the time-varying aspect of the main effect due to the action-centered term (e.g., $\pi_t$ is  continuously updated during the study).
Omitting this term would reduce the number of parameters in the model but we have found that in experiments the inclusion of $\pi_t  f(S_t)$ reduces the variance of the treatment effect estimates 
and thus speeds the learning. 
Second, in the case where the treatment effect model (\ref{model:txt effect}) is incorrect, for example, the treatment effect is non-linear in $f(S_t)$ or is time non-stationary (with time-varying $\beta$), it can be shown \cite{boruvka2018assessing} that the Bayesian regression  provides a linear approximation to the treatment effect.  When the action is not centered, the treatment effect estimates may not converge to any useful approximation at all, which could lead to poor performance in selecting the action.

The Bayesian model requires prior distributions on $\alpha_0, \alpha_1$ and $\beta$.  		Here the priors  are independent and given by 			$\alpha_0 \sim \N(\mu_{\alpha_0}, \Sigma_{\alpha_0}), ~ \alpha_1 \sim \N(\mu_{\beta}, \Sigma_{\beta}),  ~ \beta \sim \N(\mu_{\beta}, \Sigma_\beta)$; see in section \ref{sec: tuning parameters} for a discussion of how informative priors (challenge \ref{challenge: learn fast}) are constructed using HeartSteps V1 data.    Because the priors are Gaussian and the error in (\ref{model:rwrd}) is Gaussian, the  posterior distribution of $\beta$ given the current history $H_d$ is also a Gaussian, denoted by  $\N(\mu_{d+1}, \Sigma_{d+1})$.   Below we provide the details about  the calculation of $(\mu_{d+1}, \Sigma_{d+1})$.  We  first calculate the posterior distribution of all parameters, $\theta^\transpose = (\alpha_0^\transpose, \alpha_1^\transpose, \beta^\transpose )$ and the posterior distribution of $\beta$ can then be identified.  The posterior distribution of $\theta$, denoted by $\N(\bar \mu_{d+1}, \bar \Sigma_{d+1})$, given the current history $H_d = \{S_{l, k}, A_{l, k}, R_{l, k} \}_{1\leq l \leq 5, 1 \leq k \leq d}$ can be found by
\begin{align}
& \bar \Sigma_{d+1} = \left(\frac{1}{\sigma^2}\sum_{k=1}^{d}\sum_{l=1}^{5} I_{l, k}\phi(S_{l, k}, A_{l, k}) + \bar \Sigma^{-1} \right)^{-1}  \label{post var cal}\\
& \bar \mu_{d+1} = \bar \Sigma_{d+1}  \left(\frac{1}{\sigma^2}\sum_{k=1}^{d}\sum_{l=1}^{5} I_{l, k}\phi(S_{l, k}, A_{l, k})R_{l, k} + \bar \Sigma^{-1} \bar \mu\right) \label{post mean cal}
\end{align}
where $\phi(S_{l, k}, A_{l, k})^\transpose = (g(S_{l, k})^\transpose,  \pi_t f(S_{l, k})^\transpose, (A_{l, k}) - \pi_{l, k})f(S_{l, k})^\transpose )$ denotes the joint feature vector and $(\bar \mu, \bar \Sigma)$ is the prior mean and variance of $\theta$, e.g., $\bar \mu = (\mu_{\alpha_0}, \mu_{\beta}, \mu_{\beta})$ and $\bar \Sigma = \operatorname{diag}(\Sigma_{\alpha_0}, \Sigma_{\beta}, \Sigma_{\beta})$.  Suppose the size of $f(s)$ is $p$. Then the posterior mean of $\beta$, $\mu_{d+1}$ is the last $p$ elements of the above $\bar \mu_{d+1}$ and the posterior variance of $\beta$, $\Sigma_{d+1}$ is the bottom-right corner matrix of size $p$ by $p$ in $\bar \Sigma_{d+1}$.

\subsubsection{Proxy Delayed Effect on Future Rewards}
\label{sec:proxy value}

The proxy  is formed based on a simple Markov Decision Process (MDP)  for the states $S_t = (Z_t, I_t, X_t)$, in which we make the following working assumptions: 
\begin{enumerate}[label=(S\arabic*)]
	\item \label{S1} the context $\{Z_t\}$ is i.i.d. with distribution $F$, 
	\item \label{S2} the availability $\{I_t\}$ is i.i.d. with probability $p_{\avail}$
	\item \label{S3}  the dosage variable $\{X_{t}\}$ makes transitions according to $\tau(x'|x, a)$
	\item \label{S4} the mean reward given $S_t=s$ and $A_t=a$ is $r(s,a)$.  
\end{enumerate}
We use this simple MDP to capture the delayed effect on the future rewards of sending the treatment.  Note that in this model, the action only impacts the future rewards through the dosage since the context is assumed independent of the actions;  this allows us to form an estimate of delayed effect of treatment based on the current dosage.  We assume that the context and availability are both i.i.d. across time. The i.i.d. assumption leads to a reduced variance of the estimator of the delayed effect as this assumption  does not require that we have to also learn a  transition model for the context and availability.

We first discuss how each component in the simple MDP are constructed.  Given the history up to the end of day $d$, $H_d$,  we set (1) the average prior availability is  $p_{\text{avail}} = (1/5d) \sum_{k, l=1}^{d, 5} I_{l, k}$, (2) the empirical distribution on $\{Z_{l,k}\}$ is  $F(\cdot) = (1/5d) \sum_{k, l=1}^{d, 5} \delta_{Z_{l, k}}(\cdot)$ where $\delta_z(\cdot)$ is the Dirac measure, and (3) the reward function at available decision times is $r(s, a) = g(s)^\transpose \hat \alpha_0 + a f(s)^\transpose \hat \beta$ where $\hat \alpha_0$, $\hat \beta$ are the posterior means based on the model \ref{model:rwrd}. The mean reward at unavailable decision times has the same form but with posterior means from a similar linear Bayesian regression using the unavailable time points in $H_d$. 
To complete the description of the MDP, we need to specify the transition model, $\tau(x'|x, a)$ for  the dosage variable $\{X_{t}\}$. Recall that the dosage variable is defined at the beginning of section~\ref{RLf}.    Let $p_{\push}$ be the probability of delivering  any anti-sedentary suggestions between decision times given no activity suggestion was sent at the previous decision time.   We set $p_{\push} = 0.2$ based on the planned scheduling of anti-sedentary suggestions (an average of 1 anti-sedentary suggestion uniformly distributed in a 12-hour time window during the day).  Then
$\tau(x'|x, a)$ is given by  $\tau(x'|x, 1) =  \indicator{x' = \lambda x + 1}, ~ \tau(x'|x, 0) = p_{\push} \indicator{x' = \lambda x + 1} + (1-p_{\push})  \indicator{x' = \lambda x}$.   Recall from section~\ref{RLf} that $\lambda=0.95$.

We formulate the proxy of delayed effect based on the above constructed MDP as follows. Consider an arbitrary policy $\pi$ that chooses the action $\pi(S)$ at the state $S = (Z, I, X)$ if available (i.e., $I = 1$) and chooses action $0$ otherwise.   Recall the state-action value function for policy $\pi$ under discount rate $\gamma$: 
\begin{align*}
& Q^\pi(s, a)  = \EE_{\pi} [R_t+  \gamma R_{t+1} + \gamma^2 R_{t+2} + \dots |~ S_t = s, A_t = a]
\end{align*}
where the subscript $\pi$ means the actions $(A_2, A_3, \dots)$ are selected according to the policy $\pi$.     Also recall the state value function $V^\pi(s) = Q^\pi(s, \pi(s))$.  The value function $Q^\pi$ is divided into two parts: $Q^\pi(s, a) = r(s, a) + \gamma H^\pi(x, a)$ where $r(s, a)$ is the expected reward in \ref{S4} and $$H^\pi(x, a) = \EE[V^\pi(S_{t+1})|S_t=s, A_t = a] =\EE_\pi[ R_{t+1} + \gamma R_{t+2} + \gamma^2 R_{t+3} + \dots \given S_t = s, A_t = a]$$ is the sum of future discounted rewards (future value for short).  $H^\pi(x, a)$  excludes the first, immediate reward ($R_t$) and is only a function of $(x, a)$ under the working assumptions \ref{S1} and \ref{S2}. Note that the difference $H^\pi(x, 1) - H^\pi(x, 0)$ measures the impact of sending treatment at dosage $x$  on the future rewards in the setting in which future actions are selected by  policy $\pi$.   We select policy $\pi$ to maximize the future value under the constraint that $\pi$ only depends on the dosage and availability.  Specifically, let $H^*(x, a) = \max\{ H^\pi(x, a): \pi: \mathcal{X} \times \{0, 1\} \goes \A, ~\pi(x, 0) = 0, \forall x \in \mathcal{X} \}$.
It can be shown that $H^*$ is given by $H^*(x, a) = \sum_{x', i'} \tau(x'|x, a) p_{\avail}^{i'} (1-p_{\avail})^{1-i'}  V^*(x', i')$,
where the bivariate function $V^*: \mathcal{X} \times \{0, 1\} \goes \mathbb{R}$ solve the following equations:
\begin{align*}
& V(x, i) = \max_{a \in \A(i)} \bigdkh{ r_1(x, a) + \gamma \sum_{x', i'} \tau(x'|x, a) p_{\avail}^{i'} (1-p_{\avail})^{1-i'}V(x', i')} 
\end{align*}
for all $x \in \mathcal{X}$ and $i \in \{0, 1\}$, where $\A(i)$ is the constrained action space based on availability, i.e., $\A(1) = \{0, 1\}$ and $\A(0) = 0$, $r_0$ and $r_1(x, a)$ are the marginal reward function (e.g., marginal in the sense that it only depends on the dosage variable) given by $r_0(x) = \int r((z, 0, x), 0) d F(z), r_1(x, a) =  \int r((z, 1, x), a) d F(z)$. 
Finally, the proxy for the delayed effect is calculated by
\begin{align}
\eta_{d+1}(x) =  \gamma H_{d+1}(x, 0) - \gamma H_{d+1}(x, 1) \label{proxy delayed}
\end{align}
where $H_{d+1} = (1-w) H_{\init} +  w H^*$ is the weighted average between the estimate $H^*$ and the initial function $H_{\init}$ calculated based on only data from HeartSteps V1. The selection of  the discount rate $\gamma$ and the weight $w$ will be discussed in section~\ref{sec: tuning parameters}.  This delayed effect is the mean difference of the discounted future rewards between sending nothing versus an activity suggestion.  From here we see that in (\ref{action selection}) $A_t$, the action at decision time $t$ is essentially selected to maximize the sum of discounted rewards, i.e., $A_t \approx \arg \max_a \{r(S_t, a) + \gamma H_d(X_t, a)\}$.

	\subsection{Choosing Inputs}
	\label{sec: tuning parameters}
	We review the inputs required by the HeartSteps V2 RL algorithm and discuss how each is selected  based on the data collected from HeartSteps V1. The list of required inputs can be found in Figure \ref{alg}. 
	
	First, the scientific team decided $\epsilon_0 = 0.2$ and $\epsilon_1 = 0.1$ in the probability clipping to ensure enough exploration, e.g., forcing the RL algorithm continuously explore without locking into a deterministic policy.  As mentioned in section \ref{RLf}, we define the dosage in  the form of $X_{t+1} = \lambda X_{t} +  \mathds{1}_{E_{t+1}}$ (recall this variable is used to form the proxy for the delayed effect (\ref{action selection}).  Generalized Estimating Equations' (GEE)  analysis \cite{Liang1986} was conducted using HeartStep V1 data for a variety values of $\lambda$. When $\lambda$ is relatively large the dosage significantly impacts the effect of the activity suggestions on the subsequent 30 minute step count.   The scientific team selected $\lambda = 0.95$
	
	In the nightly posterior updates of treatment effect estimates, the working model (\ref{model:rwrd}) requires the features vectors, $f(s)$ and $g(s)$ (standardized to be within [0, 1]) in (\ref{model:txt effect}) and (\ref{model:baseline rwrd}) , the variance of the noise, $\sigma^2$ and the prior distribution, $\N(\mu_{\alpha_0}, \Sigma_{\alpha_0})$ and $\N(\mu_{\beta}, \Sigma_{\beta})$.  We discuss how to choose them using HeartSteps V1 data in the followings.  
	
	First, the feature vector $f(s), g(s)$ are chosen based on the GEE results using HeartSteps V1 data. In particular, each feature is included in a marginal GEE model with the prior 30-min step count in the main effect model (to reduce the variance). The feature is included in both the main effect and treatment effect model. The procedure is done for each feature separately and the $p$-value is obtained. The feature is then selected into $g(s)$ and $f(s)$ at the significance level of 0.05. 
	 For example, we found that although the 30-minute step count prior to the decision is highly predictive of the rewards (e.g., 30 minute step count after the decision), it is not significant in terms of predicting the treatment effect. Therefore, the prior 30-minute step count is  included in the baseline features $g(s)$, but not in the feature vector $f(s)$ for treatment effect.  
	A measure of how participant engages with the mobile app (e.g., the daily number of screens that participant encounters)is planned to include in both $g(s)$ and $f(s)$. This variable was not collected in HeartSteps V1. The scientific team believes this variable likely interacts with the treatment and thus decide to include into the features.  
	 The features in the feature vector $f(s)$ in (\ref{model:txt effect}) are dosage, app engagement, 
	location and the variation level of step count 60 minutes around the current time slot in past 7 days. These features along with the prior 30-minute step count, yesterday's total step count and current temperature are included in the baseline feature vector, $g(s)$.

	Second, about the variance of the noise $\sigma^2$. Although $\sigma^2$ can be learned on the fly, e.g., the residual variance by fitting the model using the data collected from the participant, to ensure the stability of the algorithm (e.g., the step count can be highly noisy), we set the variance parameter using the data from HeartSteps V1, that is, $\sigma^2$ is not updated during the study.

	Third, the prior is constructed based on the analysis result in HeartSteps V1. Specifically, we first conduct  Generalized Estimating Equations' (GEE)  regression analyses \cite{Liang1986}, using all participants' data  in HeartStep V1 and assess the significance of each feature. To form the prior variance, on each participant we fit a separate GEE linear regression model and calculated the standard deviations of the point estimates across the 37 participant models. 
	We  formed the  prior mean and prior standard deviation as follows: (1) For the features that are significant in the  GEE analysis using all participants' data, we set the prior mean to be the point estimate from this analysis; we set the prior standard deviation to the standard deviation across participant models from the participant specific GEE analyses. (2) For the features that are not significant, we set the corresponding prior mean to be zero and shrink the standard deviation by half. (3) For the app engagement variable, set the prior mean to be 0 and the standard deviation to be the average prior standard deviation of other features. 
	$\Sigma_{\alpha_0}, \Sigma_{\beta}$ are diagonal matrices  with the above prior variances on the diagonals.  The same procedure is applied to form the prior mean and variance for the reward model at the unavailable times, used in the proxy value updates.  The rationale of setting the mean to zero and shrinking the standard deviation for the non-significant features is to ensure the stability of the algorithm: unless during the HeartSteps V2 study there is strong evidence or signal detected from the participant, these features only have minimal impact on the selection of actions. In Section \ref{sec: training}, we also apply the above procedure to construct the prior in the simulation.

	The initial proxy delayed effect, $\eta_1$ and the estimates of proxy delayed effect, $\eta_{d}$ both require the  initial proxy value estimates $H_{\init}$.  To calculate $H_{\init}$ we use the same procedure as described in the section \ref{sec:proxy value}  to  calculate $H^*$,  except that the empirical probability of being available, the empirical distribution of contexts and the reward function  are constructed only using HeartSteps V1 data.

	Two remaining parameters in the HeartSteps V2 RL algorithm need to be specified: the discount rate $\gamma$ and the updating weight parameter $w$ (both part of the proxy MDP in section \ref{sec:proxy value}) For simplicity, we call them as ``tuning parameters'' in the rest of the paper. These tuning parameters are difficult to specify directly as the optimal choice likely depend on the noise level of rewards, how the context varies over time and the length of the study.  We propose to choose the tuning parameters, $(w, \gamma)$ based on a simulation-based procedure. Specifically, we first build a simulation environment (e.g., the data generating model) using HeartSteps V1 data. We then  apply the algorithm as shown in Figure \ref{alg} with each candidate pair of tuning parameters Finally, the tuning parameters is chosen such that it maximizes the total simulated rewards.  In Section \ref{sec: training}, we discuss in details how we form such generative model using HeartSteps V1 data.

	\section{Simulation Study}
	\label{sec: simulation}

In this section, we use HeartSteps V1 data to conduct a simulation study to demonstrate the validity of the procedure for choosing the inputs including the tuning parameters described in Section \ref{sec: tuning parameters}, the validity of using proxy value in the proposed algorithm addressing the challenge \ref{challenge: delay} about the negative delayed effect of treatments and the validity of using action-centering to protect against model-specification \ref{challenge: model mis and nonstationary}.   Here the use of a previous dataset to build a simulation environment for evaluating an online algorithm is similar to \cite{liao2018just}. In Section \ref{sec:real data}, we also provide the assessment of the proposed algorithm using pilot data from HeartSteps V2. 

We consider a three-fold cross validation procedure. We partition the HeartSteps V1 dataset by three folds. In each of the three iterations, two folds are marked as a \textit{training batch} and the third fold is marked as   a \textit{testing batch}.  The training batch is used to (1) construct the prior distribution, (2) form an estimate of noise variance, and (3) select the tuning parameters.  We call this process as ``training phase''.  Note that the training batch serves the same purpose as HeartSteps V1.  Next, the testing batch is used to construct a simulation environment to test the algorithm with the estimated noise variance, prior and tuning parameters.  The use of testing batch is akin to applying the RL algorithm in HeartSteps V2.  In Section \ref{sec: training} and \ref{sec:testing} below, we will describe in greater details how the training and testing batch are used \textit{in each iteration of cross validation}.  Note that we will apply the same procedure three times.

We compare the performance with \textit{Thompson Sampling Bandit} algorithm, a version similar to \cite{agrawal2013thompson}. TS Bandit algorithm is a widely used RL algorithm showing good performance in many real-world settings \cite{chapelle2011empirical}.  At each decision time, it selects the action probabilistically according to the posterior distribution of reward with the goal to maximize the immediate reward. We choose TS Bandit as the comparator over other standard contextual bandit algorithms (e.g., LinUCB in \cite{li2010contextual}) because TS Bandit is a stochastic algorithm which better suits our setting due to challenge \ref{challenge: data analysis}. Below we provide the details of TS Bandit. In TS Bandit, the expected reward is modeled by  $\EE[R_t|S_t=s, A_t = a] = r(s, a; \theta)$ for some parameter $\theta$.  At each decision time $t$ with context $S_t = s$ and availability $I_t = 1$, the action $A_t = a$ is selected with probability
$ \Pr \dkh{ r(s, a ; \theta) =  \max_{\tilde a \in \A } r(s, \tilde a ; \theta); ~ \theta \sim \N(\mu, \Sigma)}$, 
where $\N(\mu, \Sigma)$ is the posterior distribution of the parameters $\theta$ given the current history under the Bayesian model of rewards with Gaussian prior and error: $R_t = r(S_t, A_t; \theta) + \epsilon_t$. The main difference to our proposed algorithm is that TS Bandit attempts to choose action that only maximizes the immediate reward, while our proposed algorithm takes into account the longer term impact of current action for challenge \ref{challenge: delay}. In addition, the TS Bandit algorithm requires the correct modeling of each arm, while our method uses the action-centering (see (\ref{model:rwrd})) to protect against mis-specifying the baseline reward for challenge \ref{challenge: model mis and nonstationary} and only require correct modeling of the difference of two arms, i.e., the treatment effect model in (\ref{model:txt effect}).

In the implementation of TS Bandit, we parametrize the reward model by $r(s, a; \theta) = g(s)^\transpose \alpha + a f(s)^\transpose \beta$ where $f(s)$ and $g(s)$ are same feature vectors used as in our proposed algorithm.   Furthermore, to allow for a fair comparison, the prior distribution of $\theta = (\alpha, \beta)$ and the variance of error term $\sigma^2$ are both constructed by the training batch using the same procedure that will be discussed in Section \ref{sec: training} and we also clip the probability of selecting each arm with the same constraints.

\subsection{Training Phase}
\label{sec: training}

\paragraph{Prior distribution}  The algorithm requires three prior distributions: the prior of the parameters in main effect when available, the prior of parameters in the treatment effect and the prior of parameters in the mean reward when not available. The last one is used in calculating the proxy value. The prior distributions are calculated using the training batch according to Section \ref{sec: tuning parameters}

\begin{enumerate}
	\item Fit the GEE using all participants' data in the training batch (\textit{population GEE})
	\item For the parameters that are significant in \textit{population GEE}, set the prior mean to the point estimates in the \textit{population GEE}. Otherwise, set the prior mean to zero.
	
	\item For the parameters that are significant in \textit{population GEE}, the prior standard deviation is set to the standard deviation of person-specific estimates of the participants in the training batch.  Otherwise, set the prior standard deviation to  the half of the standard deviation of person-specific estimates of the participants in the training batch. 
	
	\item  The prior variance matrix of the parameters is set to the diagonal matrix. 
	
\end{enumerate}

\paragraph{Noise variance} 

Set the noise variance to be the variance of residuals obtained from the above \textit{population GEE}. 

\paragraph{Initial proxy value function}

Recall that the proxy value function requires the specification of (1) context distribution, (2) availability probability, (3) transition model of dosage and (4) reward function (for available and unavailable times), as well as the discount factor $\gamma$; see Section \ref{sec:proxy value}. We form the initial proxy using the training batch by setting (1) the empirical distribution in the training batch, (2) empirical availability probability in the training batch, (3) $p_\text{sed} = 0.2$ in the dosage transition,  and (4) the reward estimates from \textit{population GEE}.  

\paragraph{Tuning parameters}

Recall the tuning parameters are $(\gamma, w)$, corresponding to the discount rate in defining the proxy value and the updating weight in forming the estimated proxy value.  The tuning parameters are chosen to optimize the total simulated rewards using the generative model of participants in the training batch.  Below we describe how we form the generative model.  For participant $i$, we first construct a 90-day sequence of context, availability, residuals $\{Z_t^i, I_t^i,\epsilon_t^i \}_{t=1}^{450}$ as follows. 

\begin{enumerate}
	\item Create the 42-day sequence of the context, availability, residual, $\{Z_t^i, I_t^i,\epsilon_t^i\}_{t=1}^{210}$ where residual $\{\epsilon_t^i\}_{t=1}^{210}$ is obtained from the person-specific regression model fit.

	\item Extend to 90-day sequence, $\{Z_t^i, I_t^i,\epsilon_t^i\}_{t=1}^{450}$, by concatenating (90-42) days' data, randomly selected from the 42-days' data. Specifically, randomly choose $d$ from $\{1, \dots, 42\}$ and append all data from day $d$ onto the 42-day and repeat until we have a sequence of 90-day. The sampling is done only once and the sequence is fixed throughout the simulation.
		
\end{enumerate}
The generative model for participants $i$ is given as follows. At time $t  = 1, 2, 3 \dots, 450 $, 
\begin{enumerate}
	
	\item Randomly generate a binary variable $B_t$ with probability 0.2 (on average 1 per day). Here $B_t$ is the indicator of whether there is any anti-sedentary suggestion sent between $(t-1)$ and $t$.
	
	\item  Obtain the current dosage $X_t=\lambda X_{t-1}+1_{E_t}$, where $\lambda = 0.95$, the event $E_t= \{A_{t-1}=1\}\cup \{B_t=1\}$.
	
	\item Set $(Z_t, I_t) = (Z_{t}^i, I_t^i)$
	
	\item Select the action $A_t$ according to (\ref{action selection})
	
	\item Receive the reward $R_{t}$  defined as
	\begin{align}
	R_{t+1}=\left\{\begin{array}{cl}{g\left(S_{t}\right)^{\top} \alpha_{1}^{\text{train}}+A_{t} \cdot f\left(S_{t}\right)^{\top} \beta^{\text{train}}+\epsilon_{t}^i,} & {I_{t}=1} \\ {g\left(S_t\right)^{\top} \alpha_{0}^{\text{train}}+\epsilon_{t}^i,} & {I_{t}=0}\end{array}\right. \label{rwrd gen}
	\end{align}
	where the coefficients $(\alpha_{0}^{\text{train}}, \alpha_{1}^{\text{train}}, \beta^{\text{train}})$  are set based on \textit{population GEE} using all participants' data in the training batch.

\end{enumerate}

For a given candidate value of tuning parameters, together with the above constructed noise variance and prior, the algorithm is run $96$ times under each training participant's generative model. The average total reward (over training participants and re-runs) is calculated and we select the tuning parameters that maximizes the average total reward.   We use the grid search over $\gamma \in \{0, 0.25, 0.5, 0.75, 0.9, 0.95\}$ and $w \in \{0, 0.1, 0.25, 0.5, 0.75, 1\}$. Recall that the training is done three times, each time corresponding to use two folds as the training batch. The selected tuning parameters for the three iterations in CV are given by $(\gamma, w) = (0.9, 0.5), (0.9, 0.75), (0.9, 0.1)$.

\subsection{Testing Phase}

\label{sec:testing}
We build the generative model using the testing batch using the same procedure described in Section \ref{sec: training} with the only difference that in the testing phase the coefficients in generating the reward (\ref{rwrd gen}) are replaced by $(\alpha_0^{\text{test}}, \alpha_1^{\text{test}}, \beta^{\text{test}})$, which are the regression estimates using the testing dataset.  We run the algorithm under each test participant's generative model with the noise variance estimates, prior distribution and the tuning parameters selected from the training data. The algorithm is run 96 times per testing participant.  The average total reward (over re-runs) for each test participant is calculated.

\begin{figure}
	\centering
	\includegraphics[width=0.6\linewidth]{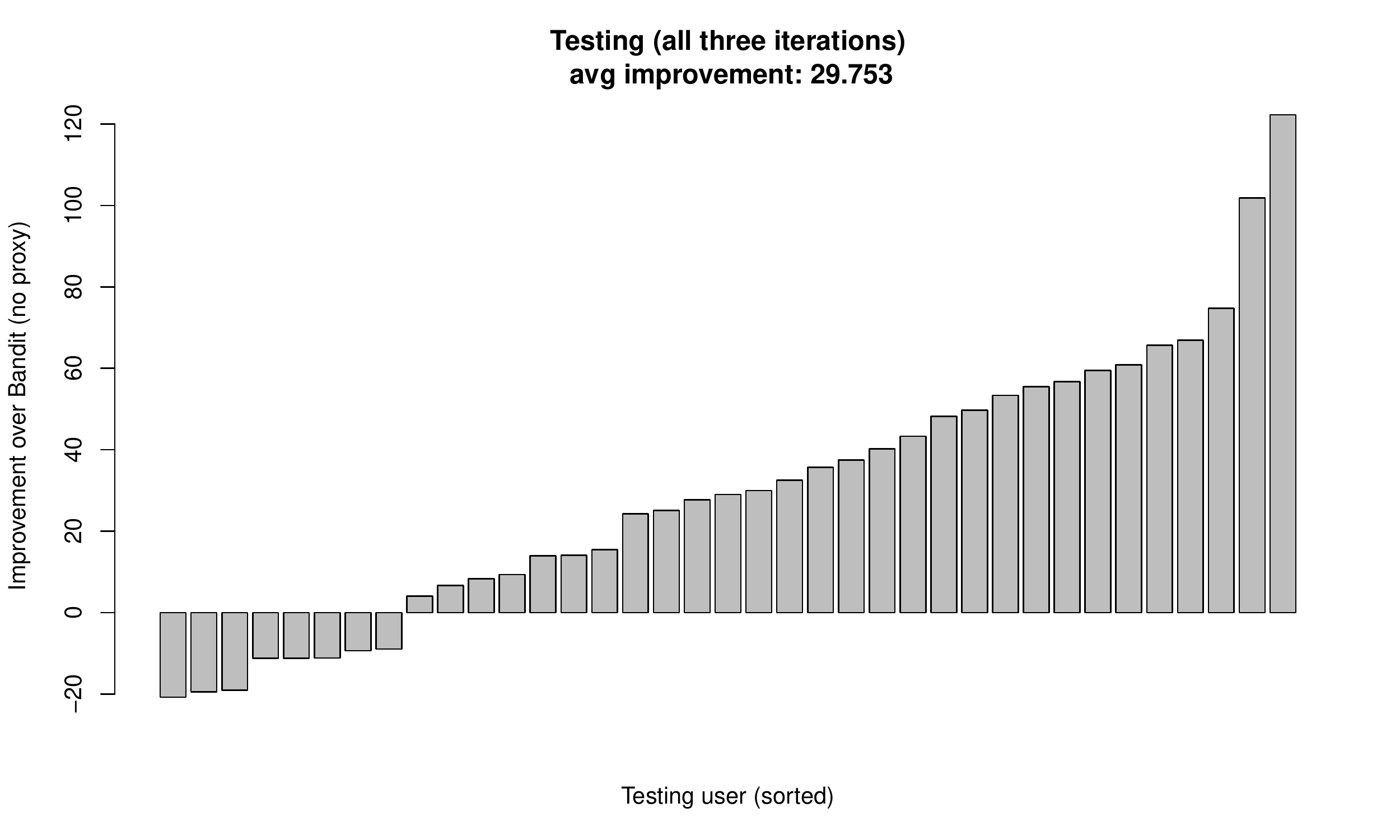} 
	\caption{Testing performance for all three iterations in the cross validation. Each bar corresponds to the improvement of total reward of the proposed algorithm with the selected inputs and tuning parameters in the training phase over the total rewards achieved by \textit{Thompson Sampling Bandit} algorithm for a single participant. }
	\label{fig:testing_all}
\end{figure}

Recall that we conduct the three-fold cross validation. Every participant in HeartSteps V1 data is assigned to a certain testing batch once in the cross validation.   The performance of the proposed algorithm and the comparator, TS Bandit algorithm, for each participant when assigned to the testing batch is provided in Figure \ref{fig:testing_all}.  We see that for the majority of participants (29 out of 37), the total rewards is higher comparing with TS Bandit algorithm. The average improvement of the total rewards over TS Bandit  is 29.753. Recall that TS Bandit algorithm is sensitive to model misspecification/non-stationarity and is greedy to maximize the immediate rewards.  The simulation results demonstrates that the use of action-centering as well as the proxy delayed effect is effective in addressing the challenge \ref{challenge: delay} and \ref{challenge: model mis and nonstationary}. 

\section{Pilot Data From HeartSteps V2 }
\label{sec:real data}
	HeartSteps V2 has been deployed in the field since June 2019. Currently the study is still in the pilot phase for testing the software and multiple intervention components.  The  RL algorithm developed above is being used to decide whether to trigger the context-tailored activity suggestion at each of the five decision times per day.  The input to the algorithm, for example the choice of feature vectors, the prior distribution and the tuning parameters, were determined according to Section \ref{sec: tuning parameters}. That is, we apply procedure described in Section \ref{sec: training} using all HeartSteps V1 data. Below we provide an initial assessment of the algorithm and also discuss the lessons learned from the pilot participants' data.

	\subsection{Initial Assessment}

	Recall that each participant in HeartSteps V2 wears the Fitbit tracker for one week prior to starting to  use the mobile app; no activity suggestion is delivered during this initial week.  Currently there are eight participants in the field who have been in the study for over one week and are experiencing the RL algorithm. For each participant, we calculate the average 30-min step count after each user-specified decision time during the first week  and compare this with the average 30-min step count in the subsequent weeks during which activity suggestion are delivered.  This is provided in Table \ref{table: pre-post steps}. All except for one participant (ID = 4) experience a positive increase in step count.  We see that on average the participant takes 125 more steps in the 30-min window following the decision time than in the first week.  
	
	\begin{table}[h]
		\centering
		\begin{tabular}{cccccc}
			\hline
			& ID & Days & \begin{tabular}[c]{@{}l@{}}Average 30-min steps \\ in the first week\end{tabular} & \begin{tabular}[c]{@{}l@{}}Average 30-min steps\\ after the first week\end{tabular} & Difference \\
			\hline
			 &   5 & 32 & 318.13 & 561.43 & 243.29 \\ 
			 &   7 & 56 & 343.79 & 574.53 & 230.75 \\ 
			 &   1 & 36& 252.12 & 424.31 & 172.19 \\ 
			 &   3 & 32 & 163.24 & 295.45 & 132.21 \\ 
			 &   8 & 18 & 281.65 & 387.86 & 106.21 \\ 
			 &   6 & 43 & 215.45 & 314.17 & 98.71 \\ 
			 &   2 & 22 & 361.26 & 418.60 & 57.35 \\ 
			 &   4 & 75 & 368.50 & 330.03 & -38.47 \\ 
			\hline
		\end{tabular}
	\caption{The average step count 30 mins after each decision time in HeartSteps V2 pilot data}
	\label{table: pre-post steps}
	\end{table}

	\subsection{Lessons}

\begin{figure}
		\centering
		\includegraphics[width=0.49\linewidth]{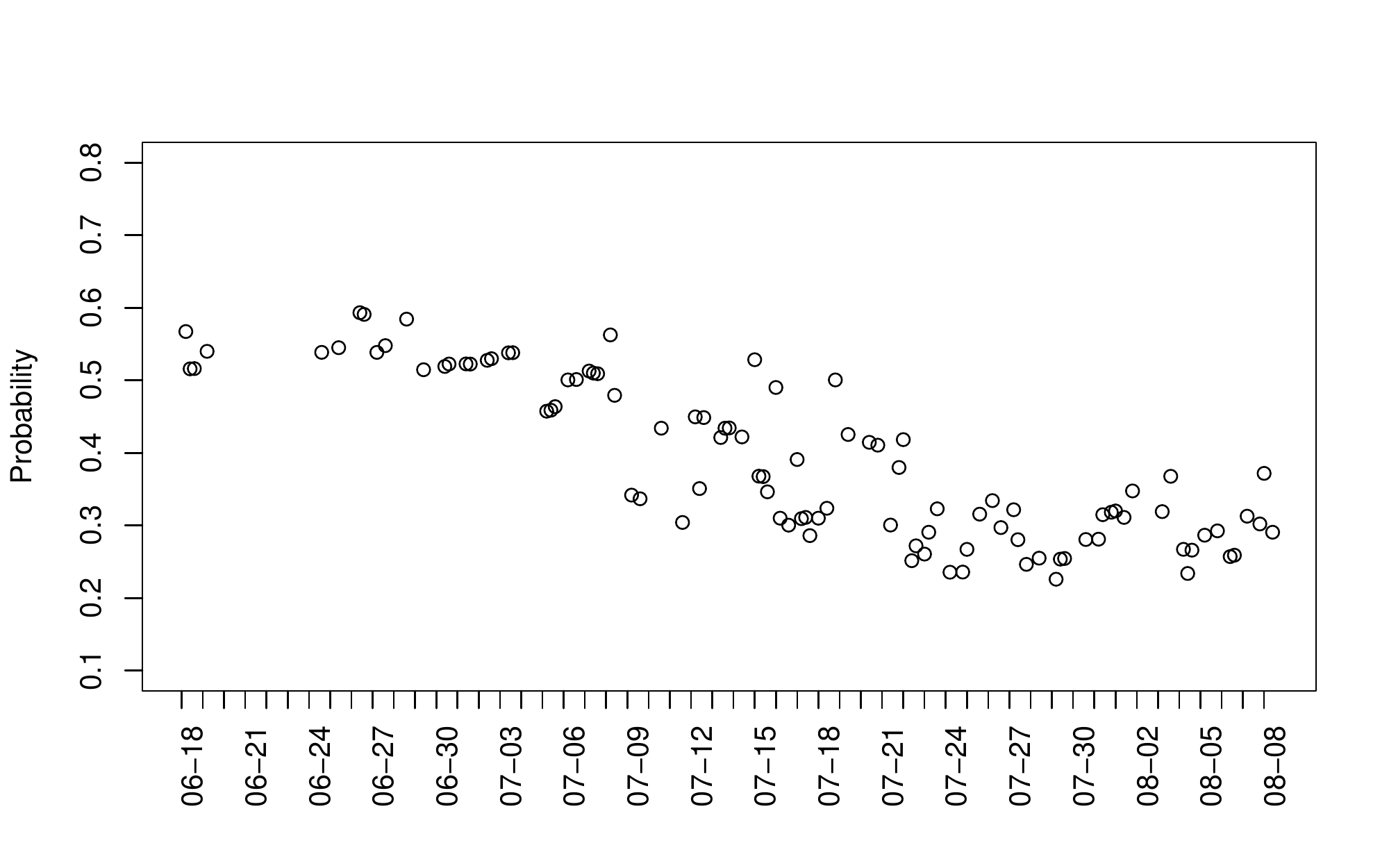}
		\includegraphics[width=0.49\linewidth]{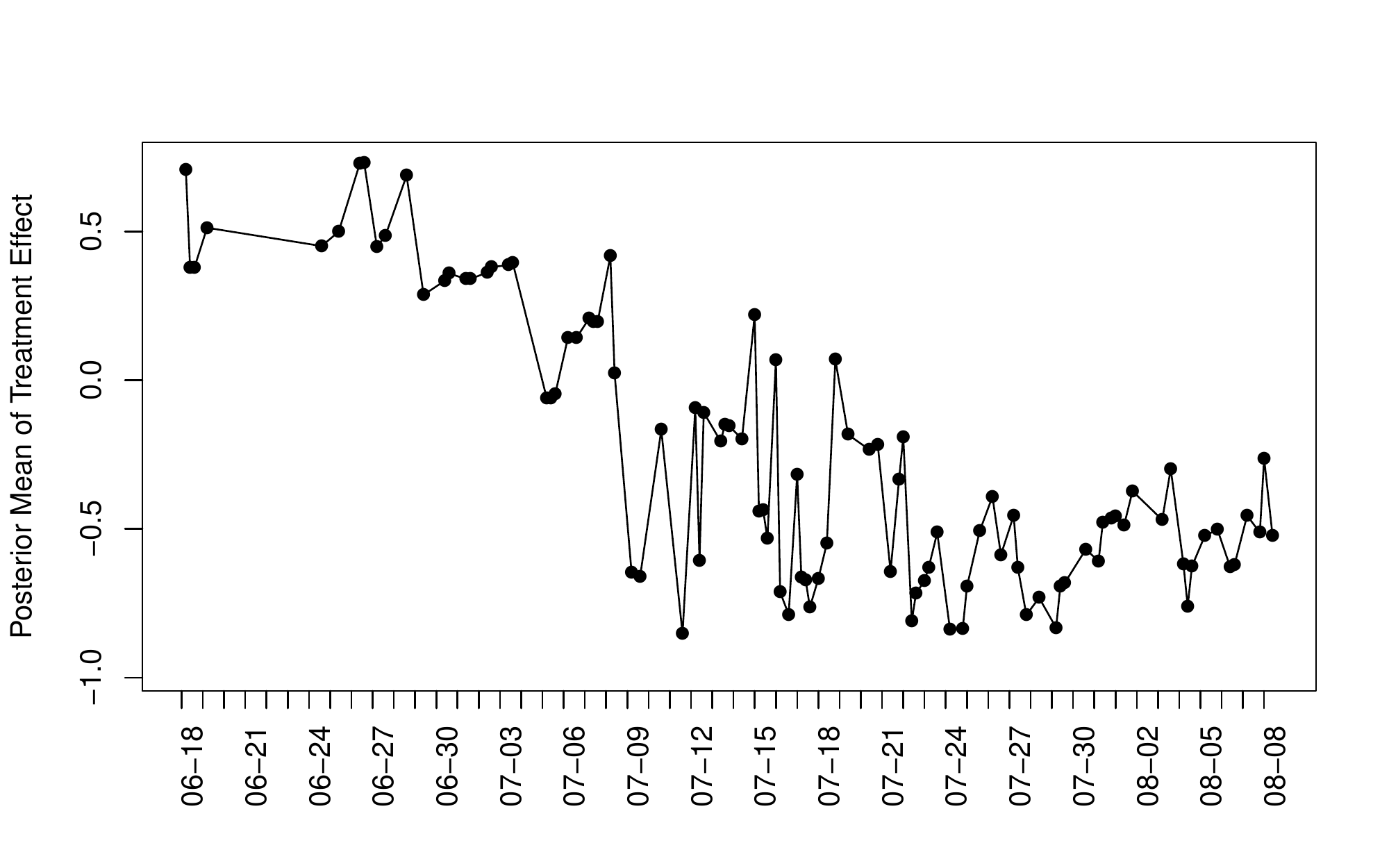}
		\caption{Participant ID = 4. \textit{Left}: the randomization probability at the available decision times.  The x-axis is the time stamp.  The y-axis is the randomization probability.  \textit{Right}: the posterior mean estimates of treatment effect at the available times. The x-axis is the time stamp.  The y-axis is the posterior mean estimates (i.e., $f(s)^\transpose \mu_d$). }
		\label{fig:103881prob}
	\end{figure}

	In this section, we discuss two lessons  learned from the examination of the pilot participants' data.  We discuss these lessons using data from participants ID=4 and ID=7.    First, consider participant ID=4 who is not responsive to the activity suggestions (i.e., sending a suggestion does not significantly improve the step count). That is, as seen in Table \ref{table: pre-post steps} participant ID = 4 has  step counts that decrease after the first week.    Figure \ref{fig:103881prob}  shows the randomization probability and the posterior mean estimates for the participant ID = 4. We see that for this participant that the posterior mean estimates start with positive value and drop below 0, e.g., no sign of the effectiveness of the suggestions, however the randomization probability still ranges between 0.2 and 0.4.   Given that HeartSteps is intended for long-term use (recall HeartSteps V2 is a 3-month study) and there are other intervention components (e.g., weekly reflection and planning and the anti-sedentary suggestion sent when participant is currently sedentary), randomizing with this probability is likely too much. 

\begin{figure}
		\centering
		\includegraphics[width=0.49\linewidth]{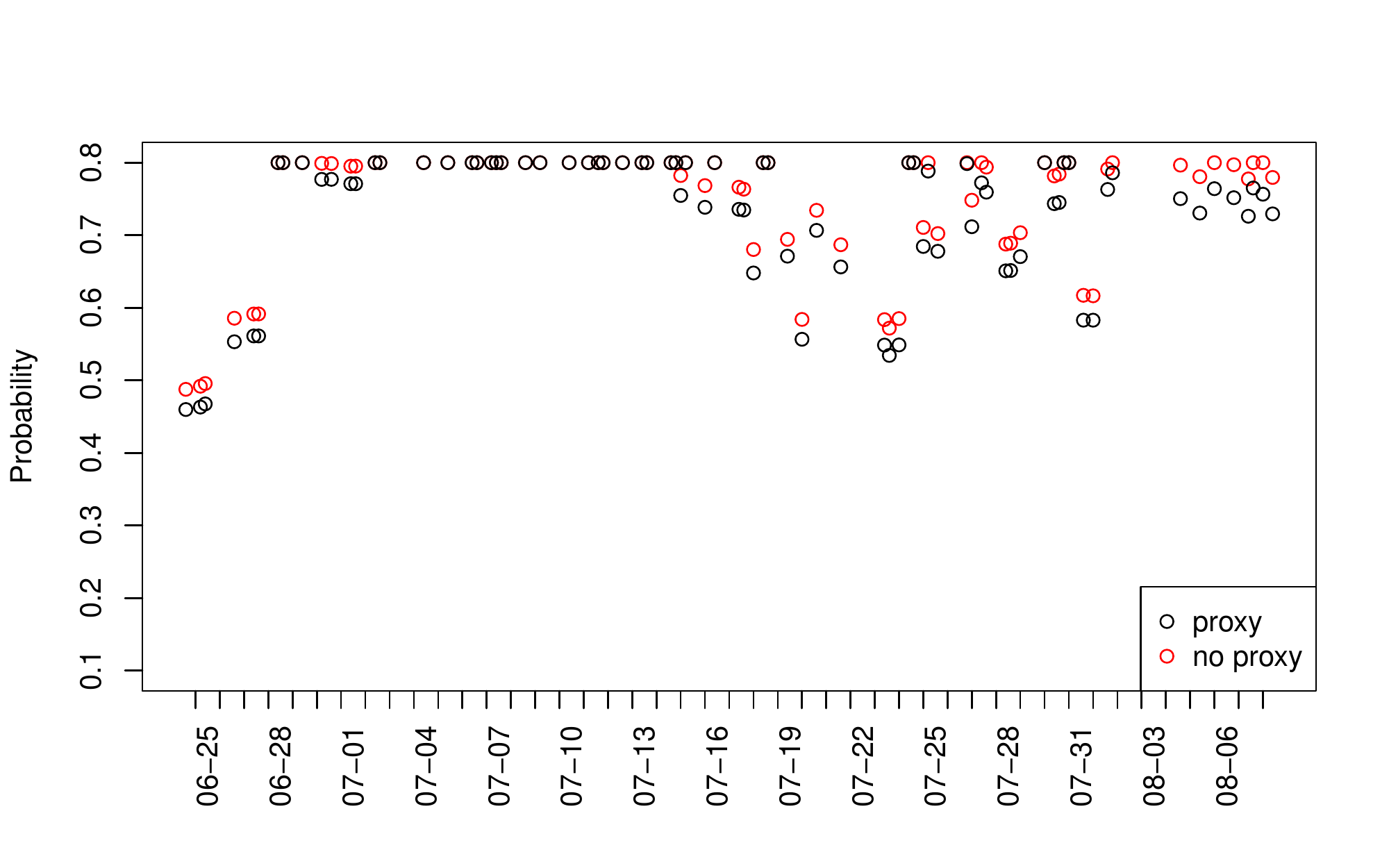} 
		\includegraphics[width=0.49\linewidth]{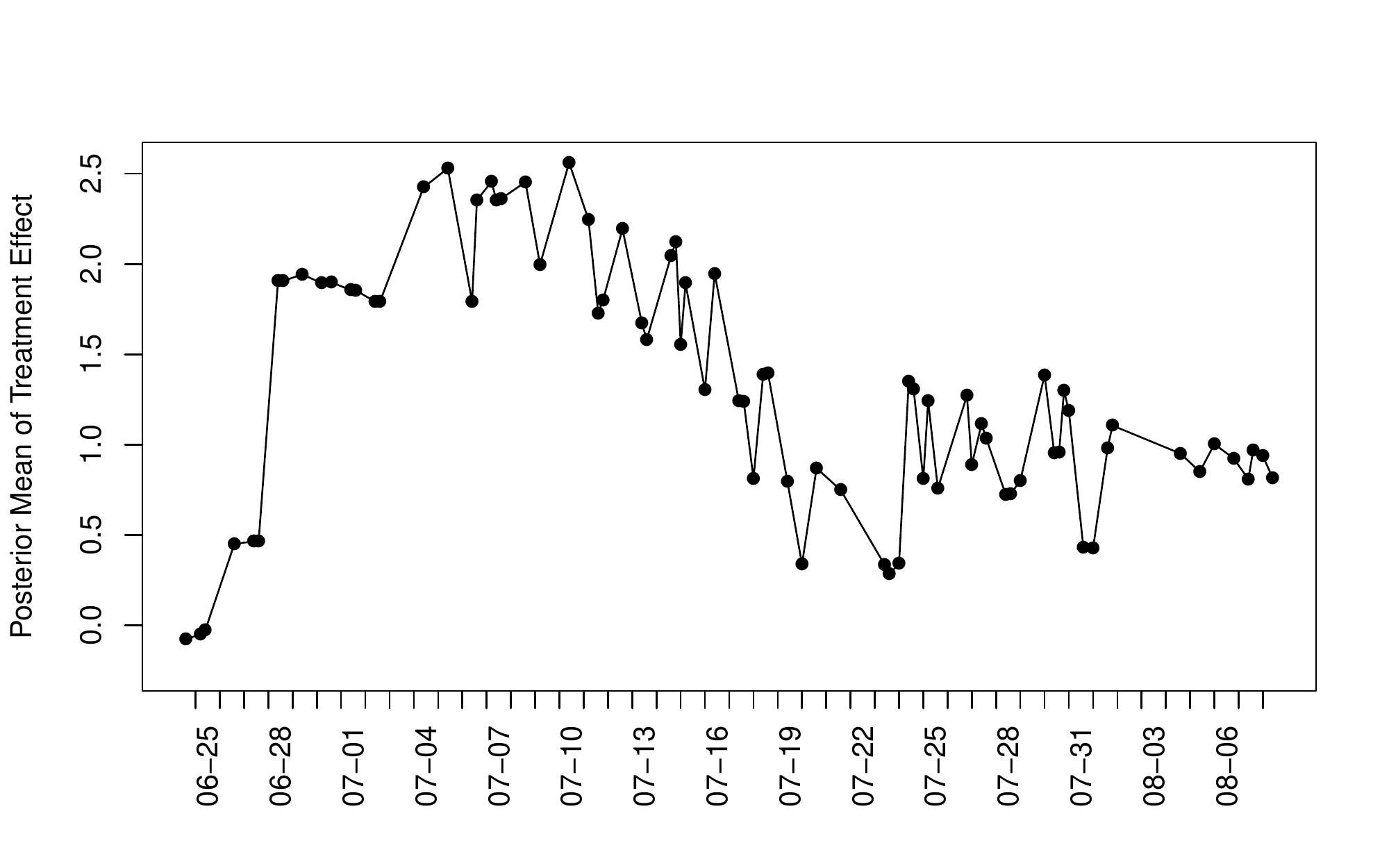}
		\caption{Participant ID = 7. \textit{Left}: the randomization probability at the available decision times.  The x-axis is the time stamp.  The y-axis is the randomization probability. The black points corresponds to the actual randomization probability and the red points corresponds to the randomization probability without the proxy  adjustment (i.e., $\eta_{d} = 0$). \textit{Right}: the posterior mean estimates of treatment effect at the available times. The x-axis is the time stamp.  The y-axis is the posterior mean estimates (i.e., $f(s)^\transpose \mu_d$). }
		\label{fig:100321prob}
	\end{figure}

	Second, consider participant ID =7 who appears highly responsive to the activity suggestions, see Table \ref{table: pre-post steps} and the right graph in Figure \ref{fig:100321prob} of the posterior mean of the treatment effect.   From the right graph in Figure \ref{fig:100321prob} we see that this participant's responsivity begins to decrease around time 07-10.  This is not that surprising as this participant is receiving many suggestions.  Next from the left graph in this same Figure the randomization probabilities from our RL algorithm does not really start to decrease until 07-16.  Ideally the proxy value should be responding quickly to the excessive dose and signalling that the probability should decrease. The  proxy value  needs improvement. The proxy is  reducing the probability of sending the walking suggestion when the delayed effect is present; see left graph in Figure \ref{fig:100321prob} and compare the black points, corresponds to the actual randomization probability with the red points corresponding to the randomization probability without the proxy  adjustment.   Ideally we would like to see a bigger gap between the black and red points in the period from 07-16 to 07-15.
	We are currently revising the algorithm in response to these two lessons; discussed in Section \ref{sec: last}
	
	\section{Conclusion and Future Work}
	\label{sec: last}

	In this paper, we developed a Reinforcement Learning algorithm for use in HeartSteps V2. Preliminary validation of the algorithm demonstrates good performance over  Thompson Sampling Bandit algorithm in  synthetic experiments constructed based on a previous study HeartSteps V1.   We also assess the performance of the algorithm using the pilot data from HeartSteps V2. After HeartSteps V2 is completed, the data will be used to further assess the performance and utility of the algorithm. 
	
	We foresee some opportunities for future work. First, our proposed algorithm learns the treatment policy separately for each participant (e.g., fully personalized). If the participants in the study are similar enough, pooling information from other participants (either currently still in the study or already having finished the study) can speed learning and achieve better performance, especially for those entering the study later. Second, the current algorithm takes into account the delayed effect of treatment by using a pre-defined ``dosage variable'' capturing the burden. It would be interesting to develop a version in which  more sophisticated measures of burden as well as engagement are used to approximate the delayed effect and  also response quicker to prevent the disengagement.  Finally, consideration of user's engagement and burden, it makes sense to reduce the chance of intervention when the algorithm does not have enough evidence of the effectiveness of intervention.

\printbibliography

\end{document}